\pdfoutput=1
\documentclass[12pt]{article}
\usepackage[utf8]{inputenc}
\usepackage{graphicx}
\usepackage[hyphens]{url}
\usepackage[]{lineno}
\usepackage{microtype}
\usepackage{booktabs}
\usepackage{authblk}
\usepackage{hyphenat}
\usepackage{subcaption}
\usepackage[style=ACM-Reference-Format]{biblatex}
\usepackage{amssymb}
\usepackage{placeins}
\usepackage[pdftex,bookmarks]{hyperref}
\usepackage{amsmath}
\usepackage{cleveref}
\usepackage{multirow}

\graphicspath{{figures/}}

\title{A Fair Experimental Comparison of Neural Network Architectures for Latent Representations of Multi-Omics for Drug Response Prediction}
\author[1]{Tony Hauptmann}
\author[1]{Stefan Kramer}
\affil[1]{Institute of Computer Science, Johannes Gutenberg University Mainz, Mainz, Germany}
\date{\today}

\begin{document}

\maketitle
%350 words
\begin{abstract}
\textbf{Background:}
Recent years have seen a surge of novel neural network architectures for the integration of multi-omics data for prediction. Most of the architectures include either encoders alone or encoders and decoders, i.e., autoencoders of various sorts, to transform multi-omics data into latent representations. One important parameter is the depth of integration: the point at which the latent representations are computed or merged, which can be either early, intermediate, or late. The literature on integration methods is growing steadily, however, close to nothing is known about the relative performance of these methods under fair experimental conditions and under consideration of different use cases.

\textbf{Results:} We developed a comparison framework that trains and optimizes  multi-omics integration methods under equal conditions. We incorporated early integration and four recently published deep learning methods: MOLI, Super.FELT, OmiEmbed, and MOMA. Further, we devised a novel method, Omics Stacking, that combines the advantages of intermediate and late integration. % Omics Stacking classifies the individual and integrated omics concurrently before it computes the final prediction with a meta learner.
Experiments were conducted on 
a public drug response data set with multiple omics data (somatic point mutations, somatic copy number profiles and gene expression profiles) that was obtained from cell lines, patient-derived xenografts, and patient samples.
Our experiments confirmed that early integration has the lowest predictive performance. Overall, architectures that integrate triplet loss achieved the best results. Statistical differences can, overall, rarely be observed, however, in terms of the average ranks of methods, Super.FELT is consistently performing best in a cross-validation setting and Omics Stacking best in an external test set setting. 

\textbf{Conclusions:} We recommend researchers to follow fair comparison protocols, as suggested in the paper. When faced with a new data set, Super.FELT is a good option in the cross-validation setting as well as Omics Stacking in the external test set setting. Statistical significances are hardly observable, despite trends in the algorithms' rankings. Future work on refined methods for transfer learning tailored for this domain may improve the situation for external test sets. The source code of all experiments is available under \url{https://github.com/kramerlab/Multi-Omics_analysis}

\end{abstract}

\section{Background}
Data analysis in the life sciences often involves the integration of data from multiple modalities or views. Integration is necessary to obtain models with improved predictive performance or explanatory power. One currently popular approach to integrating multiple views is to take advantage of latent representations as computed by neural network architectures. Views are frequently defined as (potentially very large) groups of variables that originate from one measurement technology. In bioinformatics and computational biology, views often originate from different omics platforms, e.g., from genomics, transcriptomics, proteomics, and so forth. 

The neural network architectures include either just encoders or encoders and decoders, as in autoencoder-type architectures. A myriad of architectures is possible for the integration of multi-omics data: encoders only, encoders-decoders (autoencoders of various sorts), integration either early (already for the computation of a joint latent representation), intermediate (concatenating latent representations following their computation), or late (combining the results of individual latent representations) \autocite{integrationSchemes}, and using different loss functions. As the literature on the topic is growing, one would expect that a more recent publication would improve upon a previous publication in terms of performance, or, at least, that a specific use case with superior performance has been identified then. However, as it turns out, various approaches have been optimized and tested with different sets of hyperparameters (some of them fixed, some of them optimized), with different values for hyperparameters, and with different test protocols. Further, performance can be very different for cross-validation (assuming a similar distribution of data for each test set) and so-called external test sets. Thus, it is at this point far from clear which method performs best overall, and, specifically, which method is to be preferred in which setting, e.g., in a cross-validation-like setting or with an external test set from an unknown distribution.

In this paper, we level the playing field, establish a uniform, basic protocol for various existing methods, and test the methods with a state-of-art statistical method for the comparison of machine learning algorithms \autocite{critical_difference}. As a side product, we derive a method that integrates intermediate and late integration and fares well in settings where the predictive performance on external test sets is favored. 

We study the prediction performance of the various algorithms on data sets for drug response prediction, which have been used widely in the literature in past few years \autocite{Hossein:2019}. The ultimate goal of such studies is to detect drug response biomarkers, which would help to develop personalized treatments of patients and improve clinical outcomes \autocite{Geeleher2014}. Drug screening studies on large patient cohorts are rarely performed, because it is ethically not feasible to change the chemotherapeutic regime and to cause a suboptimal therapy \autocite{Geeleher2017, Geeleher2014}.
On the other hand, large-scale drug-screening efforts using human cancer cell line models have begun to establish a collection of gene–drug associations and have uncovered potential molecular markers predictive of therapeutic responses \autocite{precision_drug_response}. A critical challenge that remains is the clinical utility of the results, i.e., the translatability from \textit{in vitro} to \textit{in vivo}  \autocite{Hossein:2019}.

% Translatability means the possibility of predicting \textit{in vivo} drug response with models trained on cell lines and was demonstrated in a previous study \autocite{Geeleher2017}. The so called translatability describes the capability of a method trained \textit{in vitro} to perform similar on \textit{in vivo} samples \autocite{Hossein:2019}.

In summary, the contributions of this paper are as follows:
\begin{itemize}
    \item a fair comparison of recent deep multi-omics integration algorithms for drug response prediction,  
    \item a new combined intermediate and late architecture for multi-omics integration, and 
    \item a thorough validation study of the methods' predictions on external test sets.
\end{itemize}

The remainder of the paper is organized as follows: First, we interpret the methods' results on the test and external sets and test significance in the observed differences. An in-depth discussion and the conclusions follow. Next, we give an overview of the used data sets and the design of the fair comparison framework. Towards the end of the article, we still give details of the included integration architectures. 
\section{Methods}
\subsection{Drug Response Data}

For our experiments, we used a publicly available drug response data set containing responses for six drugs: Docetaxel, Cisplatin, Gemcitabine, Paclitaxel, Erlotinib, and Cetuximab \autocite{Hossein:2020:dataset}. The data set was chosen, because it had \textit{patient-derived xenograft} (PDX) or patient data necessary to test the method's translatability \autocite{Hossein:2019}. Two external test sets are available for Gemcitabine.

The data set contains the drug response as target and data for three omics: somatic point mutations, somatic copy number profiles and gene expression profiles. Gene expressions are standardized and pairwise homogenized. Gene-level copy number estimates are binarized, representing  copy-neutral genes as zeroes and genes with overlapping deletions or amplifications as ones. Somatic point mutations are also in a binary format: one for mutated genes and zero for genes without a mutation \autocite{Hossein:2019}. For a fair comparison, the same data processing and training procedures were used for all methods. We used a variance threshold to filter genes with a small variance, as they hardly provide additional information. We set the variance thresholds in the same way as Park \textit{et al.}  \autocite{superfelt}. 

Sharifi-Noghabi \textit{et al.} acquired the data from  PDX mice models \autocite{pdx}, \textit{The Cancer Genome Atlas} (TCGA) \autocite{Weinstein:2013} and \textit{Genomics of Drug Sensitivity in Cancer} (GDSC) \autocite{gdsc}. GDSC consists of cell line data and was used for training, validation and testing because of its high number of samples. The trained neural networks were additionally tested on either PDX or TCGA to validate the algorithms' translatability.

The characteristics of the data set per drug are summarized in \Cref{tab:dataset_description}.  

\begin{table}[ht]
\caption{Characteristics of the drug response multi-omics data set.}
\centering
\begin{tabular}{lcccccc} 
     \toprule
     Drug & Resource &  Number of Samples &  Usage \\
     \midrule
     Cetuximab & GDSC & 856 (NR:735, R:121) &  Train \& Test \\
     Cisplatin & GDSC & 829 (NR:752, R:77) & Train \& Test\\
     Docetaxel & GDSC & 829 (NR:764, R:65) & Train \& test\\
     Erlotinib & GDSC & 362 (NR:298, R:64) & Train \& test\\
     Gemcitabine & GDSC & 844 (NR:790, R:54) & Train \& test\\
     Paclitaxel & GDSC & 389 (NR:363, R:26) & Train \& test\\
     Cetuximab & PDX & 60 (NR:55, R:5) & External test \\
     Cisplatin & TCGA & 66 (NR:6, R:60) & External test\\
     Docetaxel & TCGA & 16 (NR:8, R:8) & External test\\
     Erlotinib & PDX & 21 (NR:18, R:3) & External test\\
     Gemcitabine & PDX & 25 (NR:18, R:7) & External test\\
     Gemcitabine & TCGA & 57 (NR:36, R:21) & External test\\
     Paclitaxel & PDX & 43 (NR:38, R:5) & External test\\
     \bottomrule
     NR=Non-Responder\\
     R=Responder 
\end{tabular}
\label{tab:dataset_description}
\end{table}

\subsection{Comparison Framework}
To compare the algorithms fairly, different precautions were taken: First, the same preprocessing was performed in all experiments to provide the same input data, and second, the hyperparameters of the algorithms were optimized with an equal number of iterations by random search from a fixed grid. All algorithms draw parameters from the same grids (\Cref{tab:hyperparameter}).

\begin{table}[ht]
\caption{The hyperparameter grid used in the hyperparameter optimization.}
    \centering
    \begin{tabular}{lc}
    \toprule
         Parameter & Values \\
         \midrule
         Batch size & $\{8, 16, 32\}$\\
         Dropout rate & $\{0.1, 0.3, 0.5, 0.7\}$\\
         Epochs & $\{2..20\}$ \\
         Gamma & $\{0.0, 0.1, 0.3, 0.5\}$\\
         Layer Dimension & $\{32, 64, 128, 256, 512, 1024\}$\\
         Learning rate & $\{0.001, 0.01\}$\\
         Margin  & $\{0.2, 0.5, 1\}$ \\
         Weight Decay & $\{0.0001, 0.001, 0.01, 0.05, 0.1\}$\\
         \bottomrule
    \end{tabular}
    \label{tab:hyperparameter}
\end{table}

A $5\times5$ stratified cross-validation was performed to reduce the dependence on the data splitting. 200 hyperparameter sets were created for each iteration of the outer cross-validation and each set was used for training in the inner cross-validation. The mean \textit{area under the receiver operating characteristic} (AUROC) was used as performance measure. The algorithm was retrained with the best hyperparameter set on the combined train and validation sets. The trained network was used to compute the final results on the test sets from cross-validation and the external test set. 

The data sets are imbalanced as they contain only few responders. In the drug discovery process, researchers are mainly interested in the positive samples to find an effective drug. To account for this requirement, the \textit{area under precision recall curve} (AUPRC) \autocite{auprc} was additionally computed.

One requirement to use these methods is for them to work on patient data sets which are normally too small for transfer learning and still detect positive samples. This was covered by using the final model of an outer cross-validation iteration on an external patient or PDX data set.

The comparison of multiple algorithms on multiple data sets can lead to ambiguous results, if none of them performs significantly better. We compute the mean ranking as single value for a better comparison. Additionally, we compute the critical difference (CD) \autocite{critical_difference} with the Nemenyi significance test with $\alpha= 0.05$.

\subsection{Multi-Omics Integration Architectures}
Most deep learning algorithms for multi-omics integration are based upon the concept of encoding the feature space into a lower-dimensional latent space. The encoded representation of features in the latent space is commonly called latent features. The subnetwork that computes the latent features is called encoder and computes a non-linear dimensionality reduction. The smaller dimension of the latent representation assures that the encoder does not simply learn the identity function. An autoencoder transforms an input first into the latent representation and then decodes it back into the input dimension. The reconstructed sample should resemble the input as far as possible. The difference between input and reconstruction is measured and used as loss function  \autocite{autoencoder}.

Next, we explain briefly the six multi-omics integration architectures that were used. The first one is \textit{Early integration} (EI), which serves as baseline in our experiments. EI concatenates the omics before they serve as single input of a neural network, which consists of an encoding subnetwork and a classifier layer. The network is trained by minimizing the binary cross-entropy loss function, given that our target is to classify subjects into responders and non-responders.

\begin{figure}[ht]
    \centering
    \includegraphics[width=0.9\textwidth]{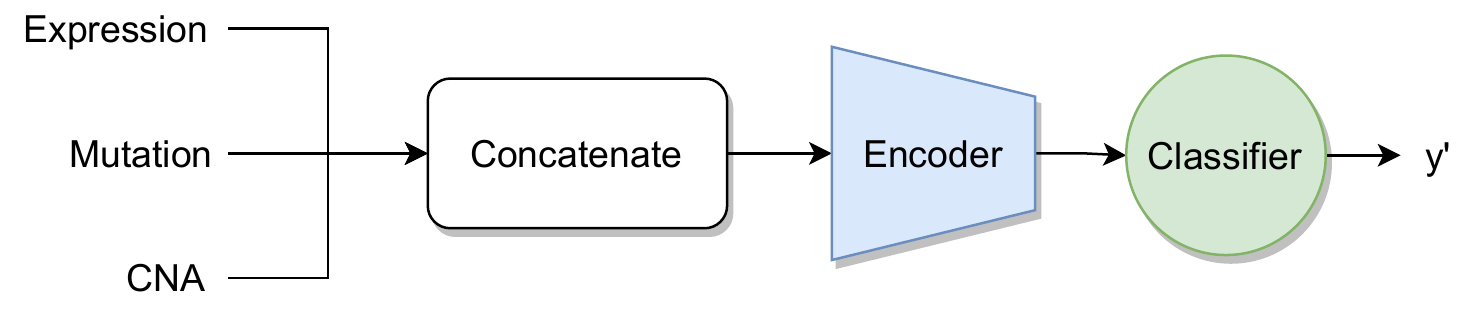}
    \caption{Schematic architecture of Early Integration with three input omics.}
    \label{fig:architecture:early_integration}
\end{figure}

The schematic architecture for three omics in visualized in \Cref{fig:architecture:early_integration}.

\begin{figure}[h]
    \centering
    \includegraphics[width=0.8\textwidth]{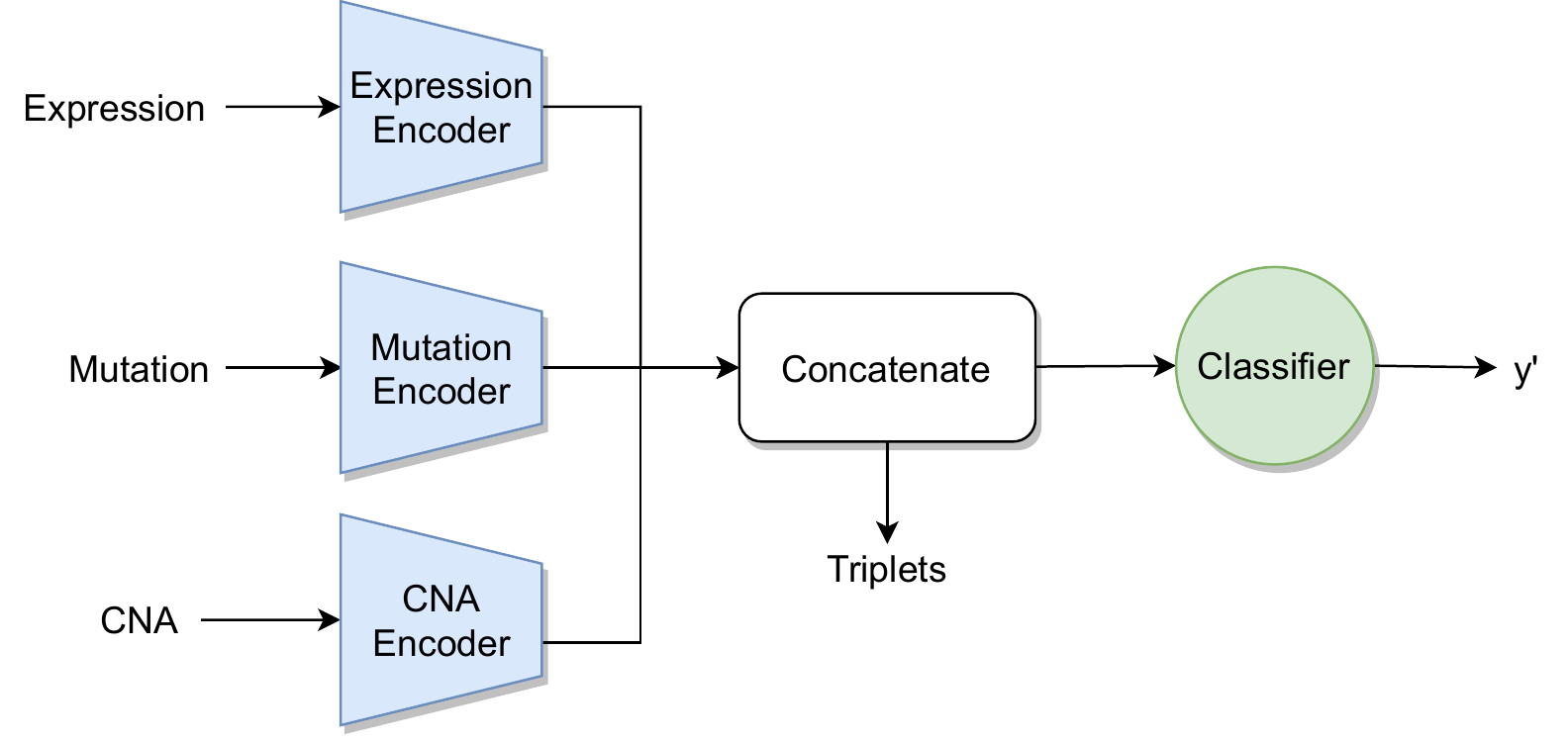}
    \caption{Schematic architecture of MOLI with three input omics.}
    \label{fig:architecture:moli}
\end{figure}

The next method is \textit{Multi-Omics Late Integration} (MOLI) developed by Sharifi-Noghabi \textit{et al.} \autocite{Hossein:2019}, which is, notwithstanding its name, an intermediate integration method. MOLI uses an individual encoder for each omics to compute the latent representations, which are concatenated and used as input for the classifier (\Cref{fig:architecture:moli}). 

Due to the small sample size, the neural network is prone to overfitting, so MOLI uses the triplet loss \autocite{triplet_loss} for regularization. The rationale behind triplet loss is that instances of the same class should have shorter distance between them than instances of different classes. To calculate the loss value, the triplet loss uses the embedding $f(x) \in \mathbb{R}^d$ of three samples. The first one is the anchor sample $x_i^a$, the second $x_i^p$ belongs to the same class, and at last a sample $x_i^n$ is of the opposite class. 

\begin{figure}[h]
    \centering
    \includegraphics[width=0.8\textwidth]{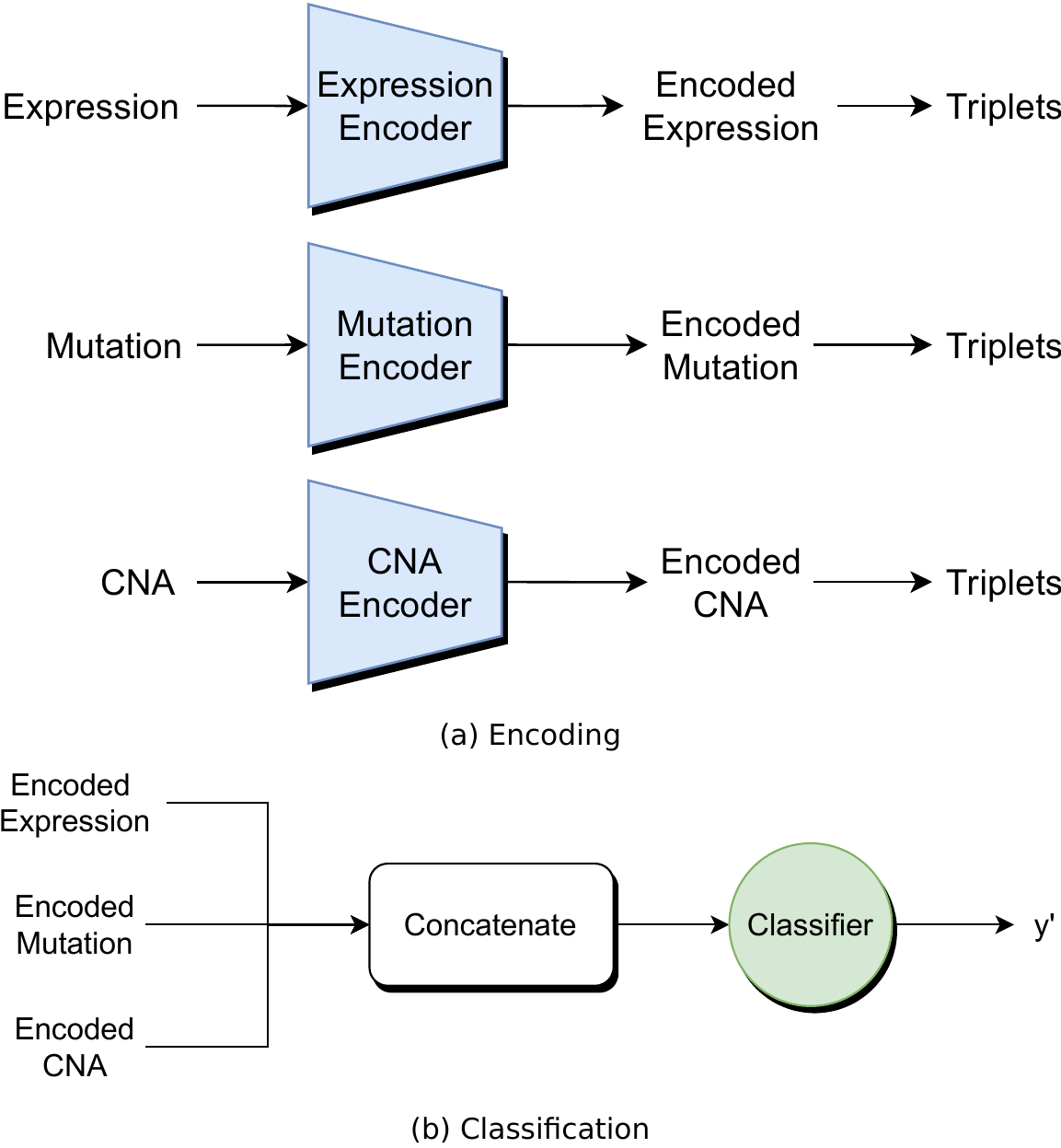}
    \caption{Schematic architecture of Super.FELT with three input omics.}
    \label{fig:architecture:super_felt}
\end{figure}

With them, the following loss function is minimized:

\begin{equation}
    \mathcal{L}_{triplet}= \sum_i^N \left[||f(x_i^a)-f(x_i^p)||_2^2- ||f(x_i^a) - f(x_i^n)||_2^2 + \alpha \right]_+,
\end{equation}

where $\alpha$ is a margin that defines the minimum distance between pairs of different classes. In MOLI the concatenated latent representations are used as embeddings. The final loss function is the sum of classification loss and triplet loss:

\begin{equation}
    \mathcal{L}_{MOLI} =  \mathcal{L}_{Classification} + \gamma  \mathcal{L}_{Triplet},
\end{equation}

where the influence of the triplet loss is weighted by the hyperparameter $\gamma$ and $\mathcal{L}_{Classification}$ is the binary cross-entropy. 
\textit{Supervised Feature Extraction Learning using Triplet Loss} (Super.FELT) is a variation of MOLI, in which the encoding and classification is not performed jointly, but in two different phases \autocite{superfelt}. The first phase trains  a supervised encoder with triplet loss for each omics to extract latent features of the high-dimensional omics data to avoid overfitting. The latent features are concatenated and used to train a classifier. \Cref{fig:architecture:super_felt} (a) shows the encoding phase and (b) the classification phase of Super.FELT.

Super.FELT added a variance threshold in the feature selection to remove features with a low variance \autocite{superfelt}. It is based on the assumption that genes with a low variance might contain less important information and can be safely removed to reduce the dimensionality, hence  increasing the focus on the more variable genes. 

Additionally, we developed a novel extension of MOLI, which we call Omics Stacking. It is inspired by  stacking and a combination of intermediate and late integration. The method stacks the results of different neural network classifier layers that use different latent features as input, with a meta-learner. The meta-learner can be any classifier, but we opted for a fully connected layer to train the neural network end-to-end.

\begin{figure}[h]
    \centering
    \includegraphics[width=0.8\textwidth]{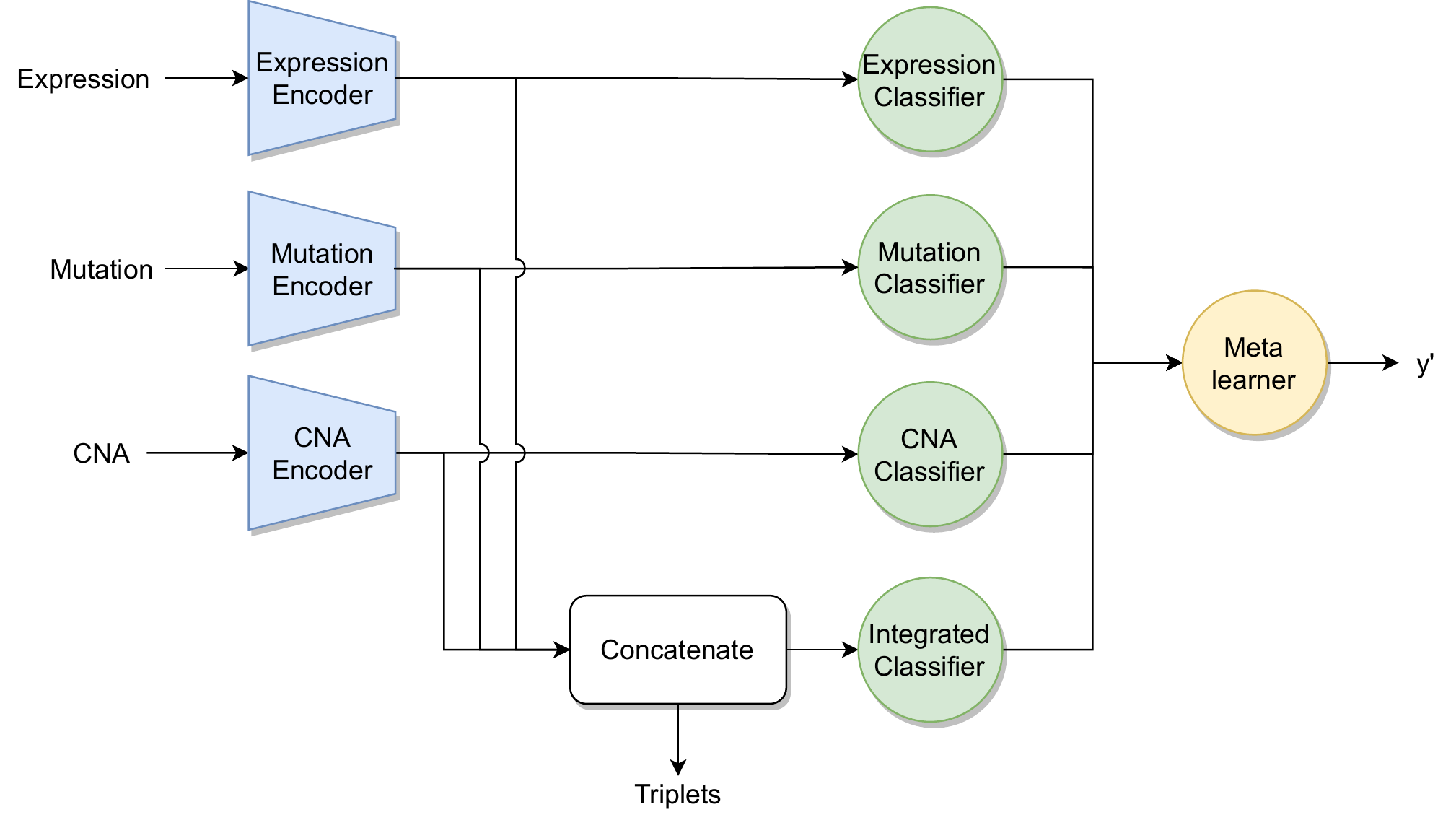}
    \caption{Schematic architecture of Omics-Stacking with three input omics.}
    \label{fig:architecture:stacking}
\end{figure}

The omics are transformed to the latent space with individual encoders, but instead of only classifying the concatenated features, Omics Stacking trains a separate classifier for each omics and the concatenated omics (\Cref{fig:architecture:stacking}). The triplet loss is still used on the concatenated embeddings to regularize. The outputs of the classifiers are combined by a meta-learner that enables the weighting of different results to emphasize the most accurate one.

Omics Stacking combines the advantages of intermediate and late integration. It models the interaction between omics and retains the weak signals of individual omics. This method relies not only on the combined omics, but also on the individual omics and so fosters generalization. We performed an ablation study to validate different versions of Omics Stacking in \Cref{ablation}.

The former methods were developed and tested on drug response prediction, but in fact can be used for other tasks as well. Next we present two methods that were validated for the prediction of a tumor's type or stage.

The first of these methods, \textit{Multi-task Attention Learning Algorithm for Multi-omics Data} (MOMA) \autocite{moma},  uses a geometrical representation and an attention mechanism \autocite{attention}. It is composed of three components: First, it builds a module for each omics using a module encoder. Each omics has its own module encoder consisting of two fully connected layers that convert omics features to modules. A module is represented by a normalized two-dimensional vector.

\begin{figure}[h]
    \centering
    \includegraphics[width=0.8\textwidth]{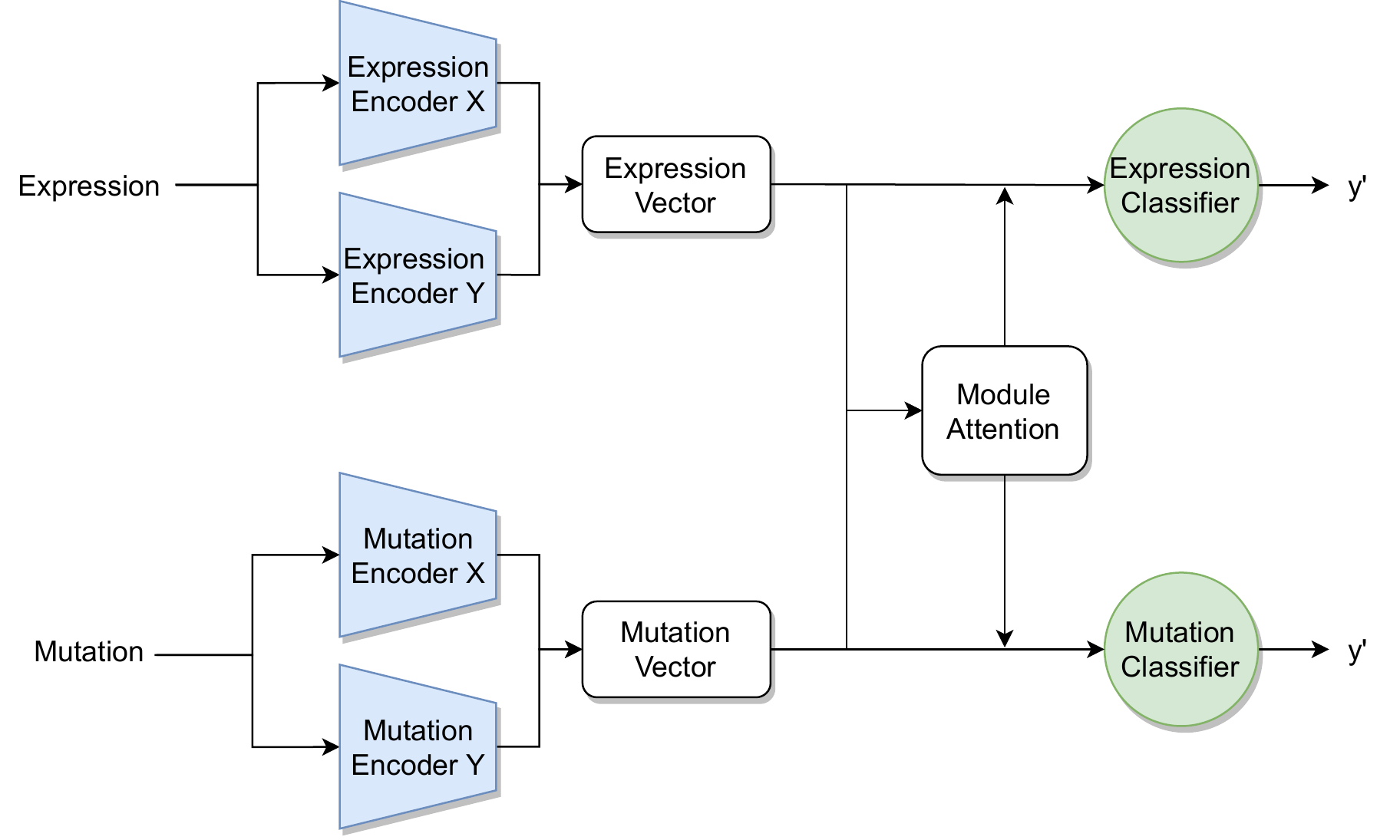}
    \caption{Schematic architecture of the network used in MOMA for two input omics.}
    \label{fig:architecture:moma}
\end{figure}

Second, it focuses on important modules between omics using a module attention mechanism. This mechanism is designed to act as a mediator to identify modules with high similarity among multiple omics. The relevance between modules is measured by the cosine similarity and is converted to a probability distribution with the softmax function. The distributions are then used to create an attention matrix that stores the relationship information between modules of different omics. To highlight important modules, the module vectors are multiplied by the attention matrix (\Cref{fig:architecture:moma}). 

Subsequently, fully connected layers and the logistic function are applied to flatten the multidimensional vectors and to compute the final probabilities for each omics \autocite{moma}.

MOMA is trained with cross-entropy loss between the true label and the omics specific outputs. After the training of the neural network, a logistic regression is fit on the omics specific outputs to generate the combined prediction.

\begin{figure}[ht]
    \centering
    \includegraphics[width=\textwidth]{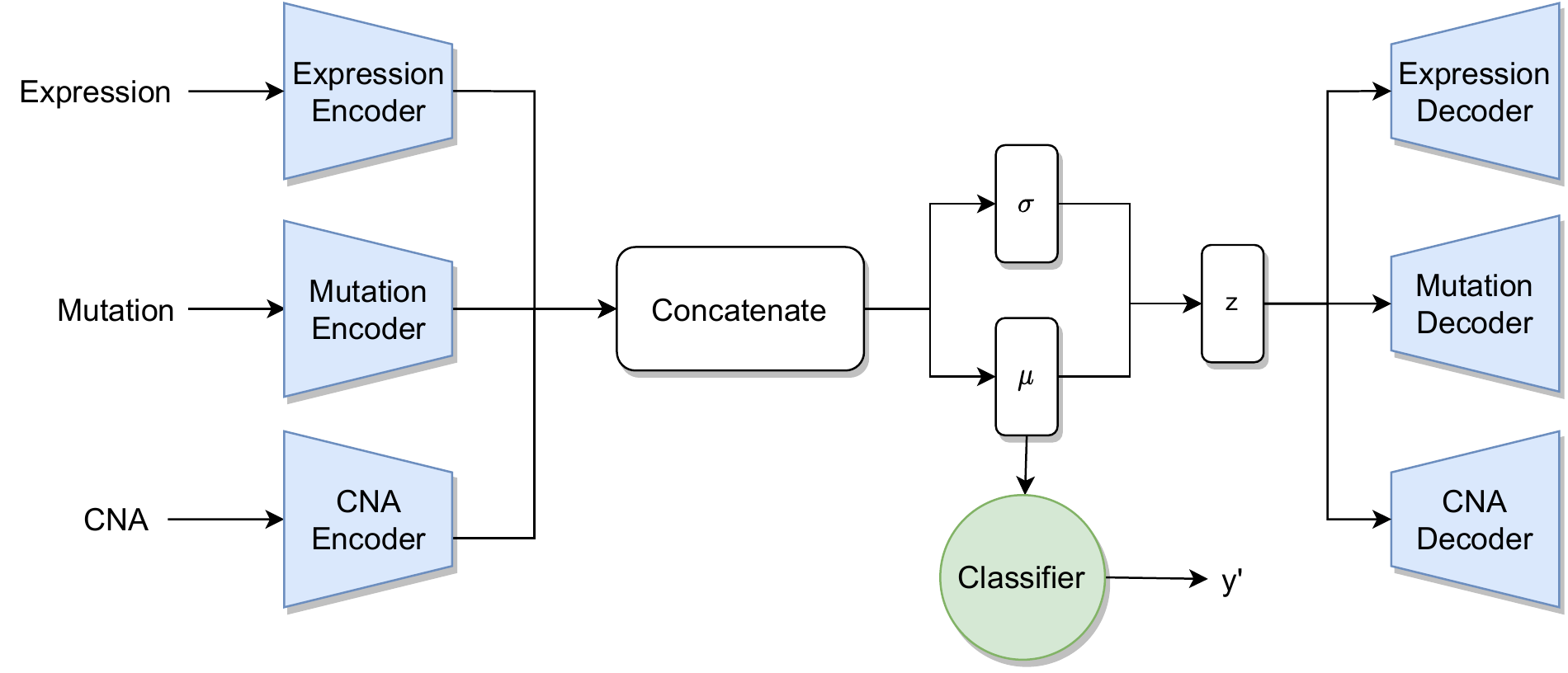}
    \caption{Schematic architecture of OmiEmbed for three input omics.}
    \label{fig:architecture:omiEmbed}
\end{figure}

The last architecture we tested OmiEmbed \autocite{omiEmbed}, which is based on a supervised variational autoencoder (VAE) \autocite{vae}. OmiEmbed was developed as a unified end-to-end multi-view multitask deep learning framework for high-dimensional multi-omics data. The method learns the latent features with the auxiliary unsupervised task of reconstructing the input omics. The trained latent features can be used for one or more supervised tasks. The overall architecture of OmiEmbed comprises a deep embedding module and one or multiple downstream task modules (\Cref{fig:architecture:omiEmbed}). 

The loss function of OmiEmbed is the sum of two parts: a reconstruction loss and the loss of the downstream task. $\mathcal{L}_{embed}$ is the unsupervised loss function of the VAE:

\begin{equation}
    \mathcal{L}_{embed} = \frac{1}{M} \sum_{i=1}^{M} BCE(x_i, x_i') + D_{KL} (\mathcal{N}(\mu, \sigma)||\mathcal{N}(0,1)).
\end{equation}

$BCE$ is the binary cross-entropy to measure the differences between the input $x$ and the reconstruction $x'$ and is computed individually for each of the $M$ omics. $D_{KL}$ is the Kullback–Leibler divergence between the learned distribution and a standard normal distribution. 

The embedding loss function is used together with the loss of the downstream task, which is in our case classification:

\begin{equation}
    \mathcal{L}_{total} = \lambda \mathcal{L}_{embed} + \mathcal{L}_{CE},
\end{equation}

where $\mathcal{L}_{CE}$ is the cross-entropy loss and $\lambda$ a balancing weight.

Instead of training all layers at the same time, the network learns in three phases: First, only the VAE is trained. In the second phase, the VAE weights are fixed and only the downstream network is trained and in the last phase the complete network is fine-tuned. 

\Cref{tab:algorithm_architecture} summarizes the components and architectures of the described methods.

\begin{table*}[ht]
    \centering
    \caption{Characteristics of the multi-omics integration methods.}
    \resizebox{\textwidth}{!}{%
    \begin{tabular}{lcccc}
    \toprule
        Architecture & Training & Triplet Loss &  Integration Type & Encoding\\
        \midrule
        Early Integration & End-to-End & - & Early & Supervised Encoder\\
        MOLI & End-to-End & + &  Intermediate & Supervised Encoder\\
        Super.FELT &  Encoding \& Classifying & + & Intermediate & Supervised Encoder\\ 
        Omics Stacking & End-to-End & + &  Intermediate + Late & Supervised Encoder\\
        MOMA & End-to-End & - & Intermediate + Late & Vector Encoding\\
        OmiEmbed & Three phases & - & Intermediate & Variational Supervised Autoencoder\\
         \bottomrule
    \end{tabular}
    }
    \label{tab:algorithm_architecture}
\end{table*}

\section{Results}
% Test set
At first, we  analyzed the results on the five cross-validation test sets (\Cref{tab:test_auroc,tab:test_auprc}). Super.FELT and Omics Stacking achieved for two drugs the highest AUROCs, but for three drugs Super.FELT was the second best and Omics Stacking for one drug. MOLI, OmiEmbed and MOMA each achieved for one drug the highest AUROC and EI not once. EI performs worse (mean rank = 5.29) than the intermediate and late integration methods. None method clearly outperforms the others, but Omics Stacking and Super.FELT performed slightly better, according to their mean ranks of 2.86 and 2.29, respectively. 

The AUPRC results are similar, but the three algorithms that use triplet loss achieved the best result for six out seven drugs. It shows the regularization benefit of triplet loss on discriminating responders and non-responders. Again, EI was the worst performing algorithm.  

All architectures have a high standard deviation for AUROC and AUPRC, which underlines the importance of stratified cross-validation to alleviate the influence of data splitting.

\begin{table*}[ht]
    \centering
    \caption{Mean AUROC on the test sets from cross-validation. Best results are shown in bold and second best are underlined. The values represent the means and standard deviations over five iterations.}
    \resizebox{\textwidth}{!}{%
    \begin{tabular}{lcccccc}
    \toprule
        Drug & Omics Stacking & MOLI & Super.FELT &  Early Integration & OmiEmbed & MOMA \\
        \midrule
        Gemcitabine TCGA & \underline{$0.646\pm0.045$} & $0.628\pm0.118$ & $0.588\pm0.070$ & $0.611\pm0.029$ & $0.628\pm0.057$ & $\mathbf{0.650\pm0.029}$ \\
        Gemcitabine PDX  & $\mathbf{0.651\pm0.071}$ & $0.622\pm0.098$ & \underline{$0.646\pm0.063$} & $0.586\pm0.092$ & $0.539\pm0.078$ & $0.625\pm0.034$\\
        Cisplatin        & $0.722\pm0.066$ & $\mathbf{0.764\pm0.039}$ & \underline{$0.753\pm0.047$} & $0.660\pm0.105$ & $0.640\pm0.071$ &  $0.714\pm0.075$\\
        Docetaxel        & $0.772\pm0.077$ & $0.792\pm0.097$ & $\mathbf{0.813\pm0.051}$ & $0.731\pm0.080$ & \underline{$0.803\pm0.035$} & $0.783\pm0.060$ \\
        Erlotinib        & $\mathbf{0.754\pm0.114}$ & $0.705\pm0.062$ & \underline{$0.744\pm0.125$} & $0.671\pm0.052$ & $0.664\pm0.130$ & $0.739\pm0.105$ \\ 
        Cetuximab        & $0.731\pm0.090$ & $0.731\pm0.033$ & $\mathbf{0.768\pm0.045}$ & $0.677\pm0.075$ & \underline{$0.754\pm0.044$} & $0.751\pm0.043$ \\
        Paclitaxel       & $0.667\pm0.138$ & $0.596\pm0.117$ & \underline{$0.726\pm0.121$} & $0.607\pm0.060$ & $\mathbf{0.740\pm0.098}$ & $0.692\pm0.081$ \\
        \midrule
        Mean Rank & \underline{2.86} & 3.86 & \textbf{2.29} & 5.29 & 3.71 & 3.00\\ 
    \bottomrule
    \end{tabular}}
    \label{tab:test_auroc}
\end{table*}

\begin{table*}[ht]
    \centering
    \caption{Mean AUPRC on the test sets from cross-validation. Best results are shown in bold and second best are underlined. The values represent the means and standard deviations over five iterations.}
    \resizebox{\textwidth}{!}{%
    \begin{tabular}{lccccccc}
    \toprule
        Drug & Omics Stacking & MOLI & Super.FELT & Early Integration & OmiEmbed & MOMA \\
        \midrule
        Gemcitabine TCGA & $\mathbf{0.161\pm0.053}$ & \underline{$0.155\pm0.065$} & $0.108\pm0.024$ & $0.117\pm0.025$ & $0.132\pm0.072$  & $0.138\pm0.041$ \\
        Gemcitabine PDX  & \underline{$0.151\pm0.085$} & $\mathbf{0.154\pm0.066}$ & $0.130\pm0.050$ & $0.138\pm0.046$ & $0.082\pm0.024$ & $0.111\pm0.027$ \\
        Cisplatin        & \underline{$0.293\pm0.089$} & $\mathbf{0.316\pm0.084}$ & $0.282\pm0.052$ & $0.222\pm0.065$ & $0.204\pm0.086$ & $0.262\pm0.075$ \\
        Docetaxel        & $0.316\pm0.083$ & \underline{$0.345\pm0.093$} & $\mathbf{0.373\pm0.027}$ & $0.312\pm0.117$ & $0.251\pm0.071$ & $0.281\pm0.090$ \\
        Erlotinib        & \underline{$0.479\pm0.153$} & $0.446\pm0.108$ & $\mathbf{0.499\pm0.162}$ & $0.346\pm0.120$ & $0.468\pm0.162$ & $0.476\pm0.151$ \\
        Cetuximab        & \underline{$0.376\pm0.104$} & $0.357\pm0.075$ & $\mathbf{0.400\pm0.095}$ & $0.290\pm0.057$ & $0.329\pm0.091$ & $0.347\pm0.070$ \\
        Paclitaxel       & $0.220\pm0.108$ & $0.163\pm0.087$ & \underline{$0.245\pm0.116$} & $0.160\pm0.045$ & $\mathbf{0.270\pm0.079}$ & $0.213\pm0.074$ \\
        \midrule
        Mean Rank & \textbf{2.14} & 2.71 & \underline{2.57} & 5.0 & 4.57 & 4.00 \\ 
        \bottomrule
    \end{tabular}}
    \label{tab:test_auprc}
\end{table*}

\begin{figure}[ht]
        \centering
        \includegraphics[width=0.8\textwidth]{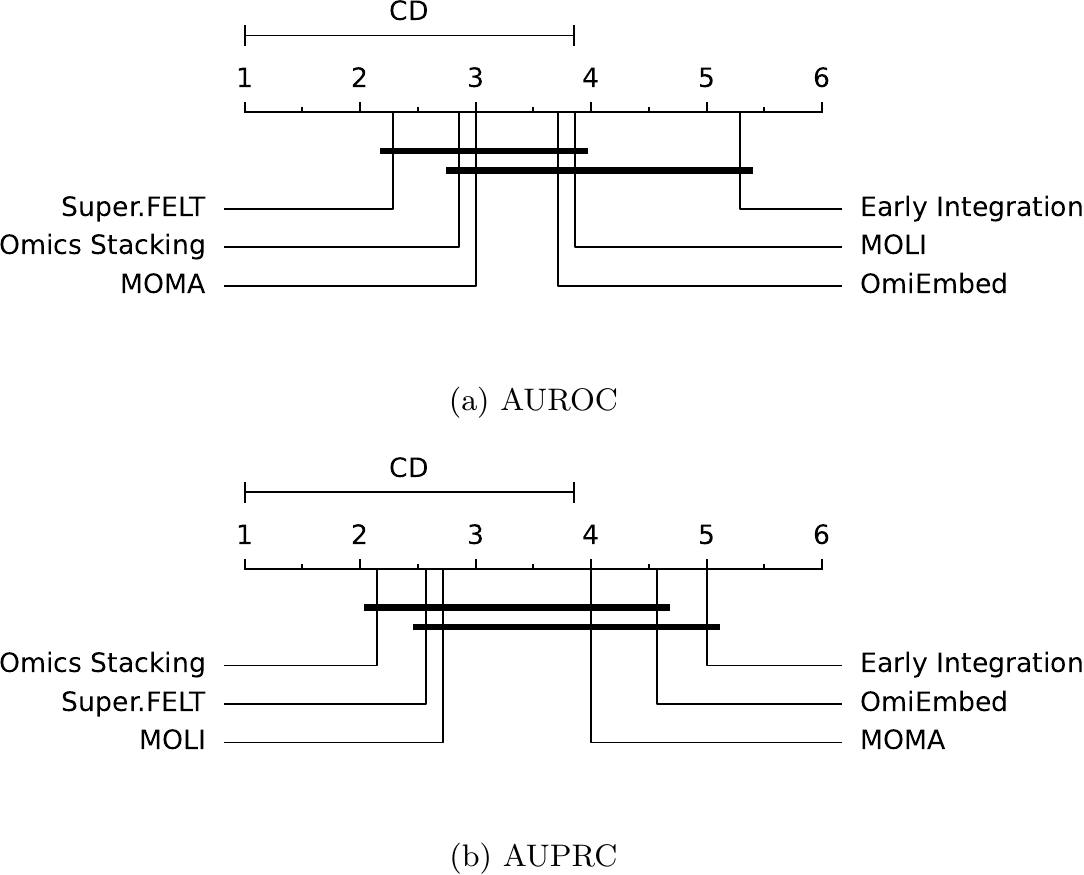}
        \caption{Mean rank and critical difference of the AUROC and AUPRC on the test sets from cross-validation. The mean values of the outer cross-validation results are compared. The Nemenyi test with $\alpha=0.05 $ was used to compute significant differences.}
        \label{fig:critical_difference_test}
\end{figure}

The visualization of the mean ranks and the critical differences   (\Cref{fig:critical_difference_test}) supports our analysis. EI is, for AUROC as well as AUPRC, significantly worse than the best performing method. The mean ranks for AUPRC contain a visible gap between methods that use triplet loss (Omics Stacking, Super.FELT and MOLI) and methods without (MOMA, OmiEmbed and EI), but without significance.

% External data
Next, we analyzed the results on the external test sets (\Cref{tab:extern_auroc,tab:extern_auprc}). Here, the method learns the source distribution, but at the same time, must generalize enough to predict drug response on an unknown distribution. 

\begin{figure}[h]
        \centering
        \includegraphics[width=0.8\textwidth]{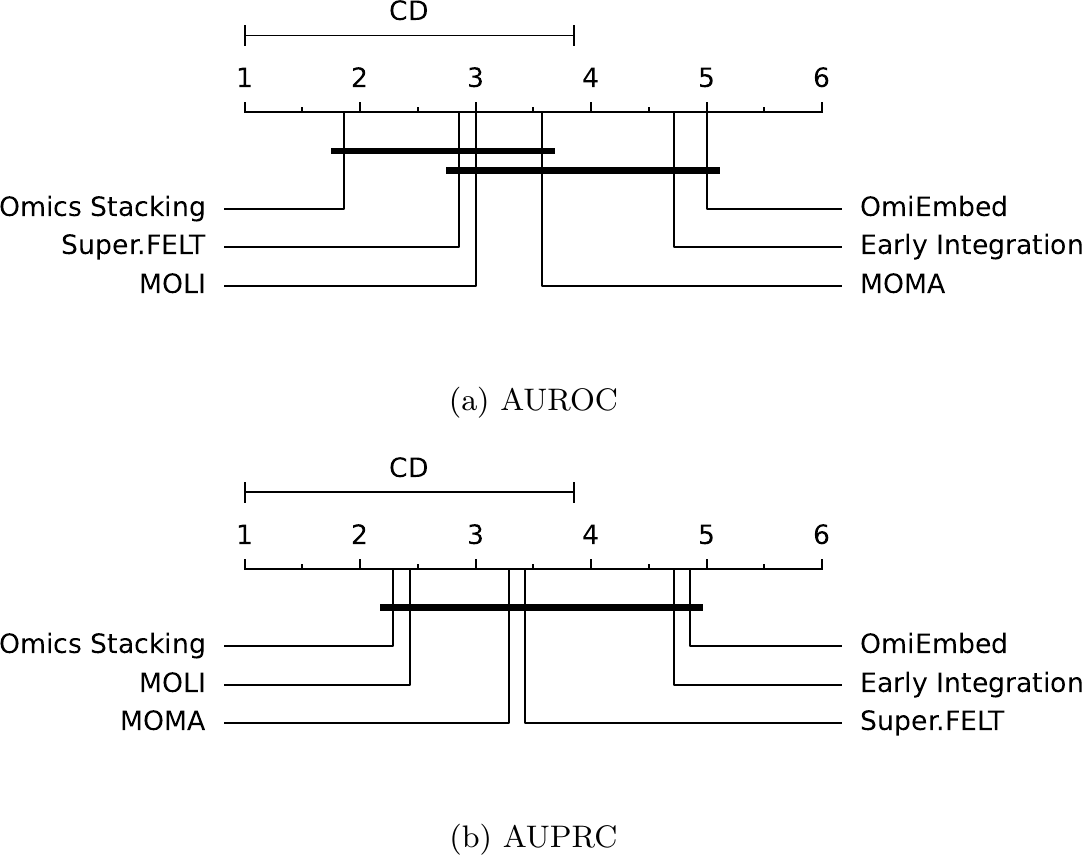}
        \caption{Mean Rank and critical difference of the AUROC and AUPRC on the external test set. The mean values of the outer cross-validation results are compared. The Nemenyi test with $\alpha=0.05 $ was used to compute significant differences.}
        \label{fig:critical_difference_external}
\end{figure}

Again, no single method performed best with all drugs, but Omics Stacking had the lowest mean rank in both metrics. It achieved the best results in half of the data sets for AUROC and AUPRC. Additionally, it achieved the lowest mean rank (see \Cref{fig:critical_difference_external} for the critical differences) for AUROC and AUPRC. The results validate the benefit of classifying both individual and integrated features for the translatability. EI achieved the best AUROC and second best AUPRC for Erlotinib, but classified worse in general. Surprisingly, OmiEmbed performed worst on the external data. One possible explanation may be that the regularization of the VAE narrows its capabilities to perform similarly on a shifted distribution without retraining. 

Differences occurred between AUROC and AUPRC: Super.FELT had the second best mean rank for AUROC, but was only fourth for AUPRC. No method was significantly better regarding the AUPRC, however, OmiEmbed and EI had similarly low mean ranks. Omics Stacking and MOLI performed similarly on the AUPRC, but Omics Stacking had higher AUROC values.

\begin{table*}[ht]
    \centering
    \caption{Mean AUROC on external test set. Best results are shown in bold and second best are underlined. The values represent the means and standard deviations over five iterations.}
    \resizebox{\textwidth}{!}{%
    \begin{tabular}{lcccccccc}
    \toprule
        Drug & Omics Stacking & MOLI & Super.FELT & Early Integration & OmiEmbed & MOMA \\
        \midrule
        Gemcitabine TCGA &  $\mathbf{0.655\pm0.029}$ & \underline{$0.640\pm0.037$} & $0.618\pm0.042$ & $0.604\pm0.093$  & $0.565\pm0.059$ & $0.473\pm0.039$ \\
        Gemcitabine PDX  &  $\mathbf{0.714\pm0.089}$ & $0.614\pm0.044$ & \underline{$0.692\pm0.054$} & $0.525\pm0.099$ & $0.657\pm0.119$ & $0.627\pm0.092$ \\
        Cisplatin        &  $0.644\pm0.087$ & $0.674\pm0.032$ & $\mathbf{0.728\pm0.045}$ & $0.604\pm0.052$ & $0.513\pm0.056$ & \underline{$0.687\pm0.019$} \\
        Docetaxel        &  $0.584\pm0.101$ & $\mathbf{0.647\pm0.038}$ & \underline{$0.588\pm0.056$} & $0.456\pm0.065$ & $0.478\pm0.056$ & $0.581\pm0.064$ \\
        Erlotinib        &  \underline{$0.744\pm0.065$} & $0.722\pm0.127$ & $0.563\pm0.080$ & $\mathbf{0.789\pm0.079}$ & $0.633\pm0.105$ & $0.715\pm0.132$ \\
        Cetuximab        &  $\mathbf{0.575\pm0.049}$ & $0.476\pm0.111$ & \underline{$0.556\pm0.099$} & $0.470\pm0.130$ & $0.468\pm0.075$ & $0.505\pm0.019$ \\
        Paclitaxel       &  $\mathbf{0.619\pm0.152}$ & $0.547\pm0.121$ & $0.527\pm0.114$ & $0.418\pm0.062$ & $0.516\pm0.026$ & \underline{$0.573\pm0.124$} \\
        \midrule
        Mean Rank & \textbf{1.86} & 3.00 & \underline{2.86} & 4.71 & 5.00 & 3.57   \\ 
    \bottomrule
    \end{tabular}}
    \label{tab:extern_auroc}
\end{table*}

\begin{table*}[ht]
    \centering
    \caption{Mean AUPRC on external test set. Best results are shown in bold and second best are underlined. The values represent the means and standard deviations over five iterations.}
    \resizebox{\textwidth}{!}{%
    \begin{tabular}{lcccccccc}
    \toprule
        Drug & Omics Stacking & MOLI & Super.FELT & Early Integration & OmiEmbed & MOMA \\
        \midrule
        Gemcitabine TCGA & $\mathbf{0.581\pm0.074}$ & \underline{$0.535\pm0.085$} & $0.502\pm0.082$ & $0.513\pm0.086$ & $0.462\pm0.095$ & $0.389\pm0.040$ \\
        Gemcitabine PDX  & $\mathbf{0.510\pm0.132}$ & $0.424\pm0.038$ & $0.457\pm0.055$ & $0.362\pm0.108$ & \underline{$0.466\pm0.100$} & $0.414\pm0.068$ \\
        Cisplatin        & $0.942\pm0.027$ & $0.950\pm0.007$ & $\mathbf{0.963\pm0.009}$ & $0.932\pm0.004$ & $0.908\pm0.009$ & \underline{$0.952\pm0.006$} \\
        Docetaxel        & $0.560\pm0.051$ & $\mathbf{0.590\pm0.021}$ & $0.565\pm0.024$ & $0.491\pm0.034$ & $0.544\pm0.049$ & \underline{$0.578\pm0.060$} \\
        Erlotinib        & $\mathbf{0.440\pm0.075}$ & $0.410\pm0.158$ & $0.223\pm0.024$ & \underline{$0.428\pm0.106$} & $0.294\pm0.079$ & $0.369\pm0.146$ \\
        Cetuximab        & $0.125\pm0.018$ & $\underline{0.141\pm0.086}$ & $0.126\pm0.039$ & $0.108\pm0.028$ & $0.101\pm0.018$ & $\mathbf{0.148\pm0.065}$ \\
        Paclitaxel       & $\mathbf{0.256\pm0.147}$ & \underline{$0.191\pm0.077$} & $0.147\pm0.028$ & $0.120\pm0.016$ & $0.135\pm0.007$ & $0.172\pm0.047$ \\
        \midrule
        Mean Rank & \textbf{2.29} & \underline{2.43} & 3.43 & 4.71 & 4.86 & 3.29 \\  
    \bottomrule
    \end{tabular}}
    \label{tab:extern_auprc}
\end{table*}

\section{Discussion}
The emerging interest in multi-omics integration with neural networks produces an increasing number of different architectures. In this work, we compared a subset of recently published methods as fairly as possible to validate their predictive capabilities. Our experiments focused on drug response prediction, a high-dimensional problem, with the added complexity of having few samples.

Our work includes only data sets from cancer research, and there are two reasons for that decision. The first one is the availability of multi-omics data, as cancer samples are more common compared to other diseases. Data sets for other diseases with enough samples and a high enough quality are still rare. The second reason is that the extensive hyperparameter optimization with an inner and outer cross-validation increases the hardware requirements. The sheer amount of different, recently published algorithms made it necessary to focus on just one area.

Current drug response methods are trained on cell lines and predict directly on patient samples without adaptation. Transfer learning is a method where
a model is trained on source data sets and later transferred to a target data set \autocite{transfer_learning}. One promising current direction is to train on \textit{in vitro} samples and fine-tune the neural network for \textit{in vivo} samples. 

Previous studies \autocite{Hossein:2019} showed that EI is not suitable for multi-omics integration, because concatenating the features leads to a higher-dimensional sparse space. Our experiments corroborate this hypothesis, because EI achieved the worst results on both data sets. This make early integration not suitable as the sole baseline: At least one other intermediate integration method should be included. However, our experiments showed that no method performs the best for all drugs.
\section{Conclusions}
In this paper we showed that none of the current multi-omics integration methods excels at drug response prediction, however, Early Integration performed significantly worse.
Researchers should not rely on a single method, but rather consider more than one method for their task at hand. If the number of experiments is limited or translatability is wanted, we recommend to use the newly introduced method, Omics Stacking, as it achieved good results on test and external data. When faced with a new data set in a cross-validation like setting, Super.FELT is another good option. Our experiments also suggest that a fair experimental is necessary to see the strengths and weaknesses of various algorithms, which were not visible from the publications alone. We hope that this comparison has shed some light on the relative performance of multi-omics integration methods, has produced valuable insights for their application, and that it encourages further research.
\FloatBarrier

\section*{Declarations}
\subsection*{Acknowledgment}

\subsection*{Funding}
This work was funded by the German Federal Ministry for Education and Research as part of the DIASyM project under grant number [031L0217A].

\subsection*{Contribution}
TH implemented the methods and experiments and wrote the manuscript. \\
SK supervised the study, developed the concept and wrote and previewed the manuscript.

\printbibliography 

\appendix
\section{Implementation Details}
The algorithms were implemented in PyTorch 1.11.0 and NumPy 1.22.4 was used to compute the AUROC and AUPRC. The critical differences were computed and visualized with Orange3 3.32.0. The triplets were generated online with an all-triplets scheme, which creates all possible triplets of a batch, as in the experiments of the corresponding paper of MOLI and Super.FELT. An open source implementation of triplet loss and triplet creation was used\footnote{https://github.com/adambielski/siamese-triplet}. Adagrad \cite{adagrad} was used to update the neural network weights.
 
Early Integration and Omics Stacking were implemented by the authors. For MOMA we used the provided example source code\footnote{https://github.com/DMCB-GIST/MOMA} and adapted it to three omics, made the number of modules a hyperparameter and combined the individual probability outputs with a logistic regression. For omiEmbed\footnote{https://github.com/zhangxiaoyu11/OmiEmbed}, MOLI\footnote{https://github.com/hosseinshn/MOLI} and Super.FELT\footnote{https://github.com/DMCB-GIST/Super.FELT}, the published source code was used.

The inner cross-validation was stopped early, if achieving a higher AUROC than the best result was not possible. This allowed us to speed up the hyperparameter optimization.

 \section{Ablation Study} \label{ablation}
In the ablation study the impact of the components, especially the influence of different numbers of concurrent classification layers, was validated. The first altered architecture -- Omics Stacking without integration -- omits the subnetwork that classifies the concatenated omics features. That alteration result in a late integration neural network. The second one -- Omics Stacking Complete -- has classifier layers for every possible non-empty subset of the omics set, which adds three additional classification subnetworks. At last, we tested Omics Stacking without Triplet Loss, which leaves out the triplet loss function. 

\begin{figure}[h]
\centering
        \includegraphics[width=0.8\textwidth]{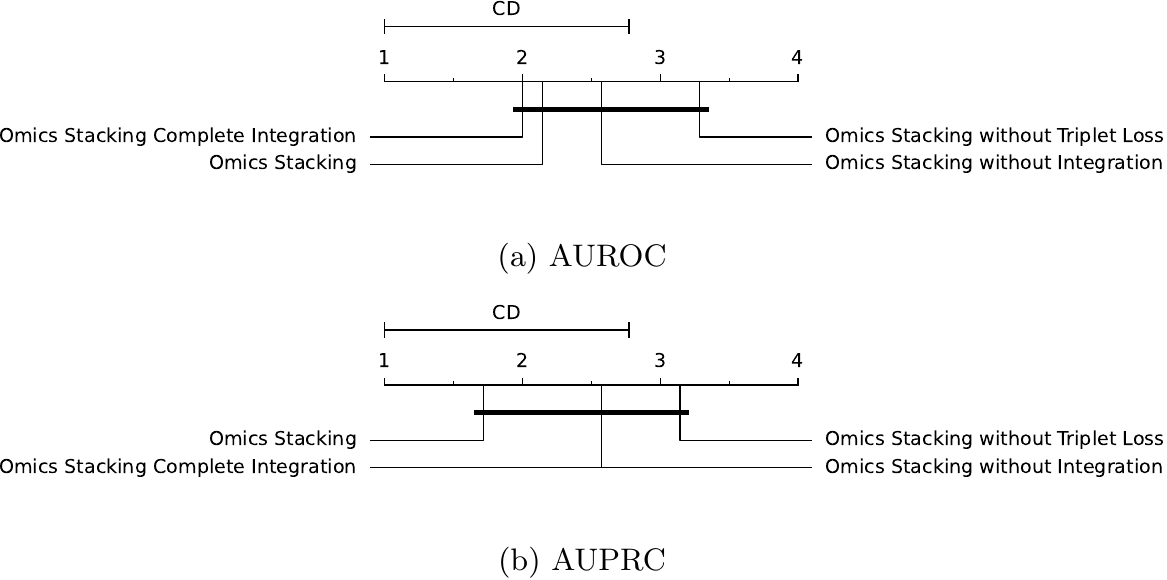}
        \caption{Mean rank and critical difference of the AUROC and AUPRC on the test sets from cross-validation for the ablation study. The mean values of the outer cross-validation results are compared. The Nemenyi test with $\alpha=0.05 $ was used to compute significant differences.}
        \label{fig:critical_difference:test_ablation}
\end{figure}

\begin{figure}[h]
\centering
        \includegraphics[width=0.8\textwidth]{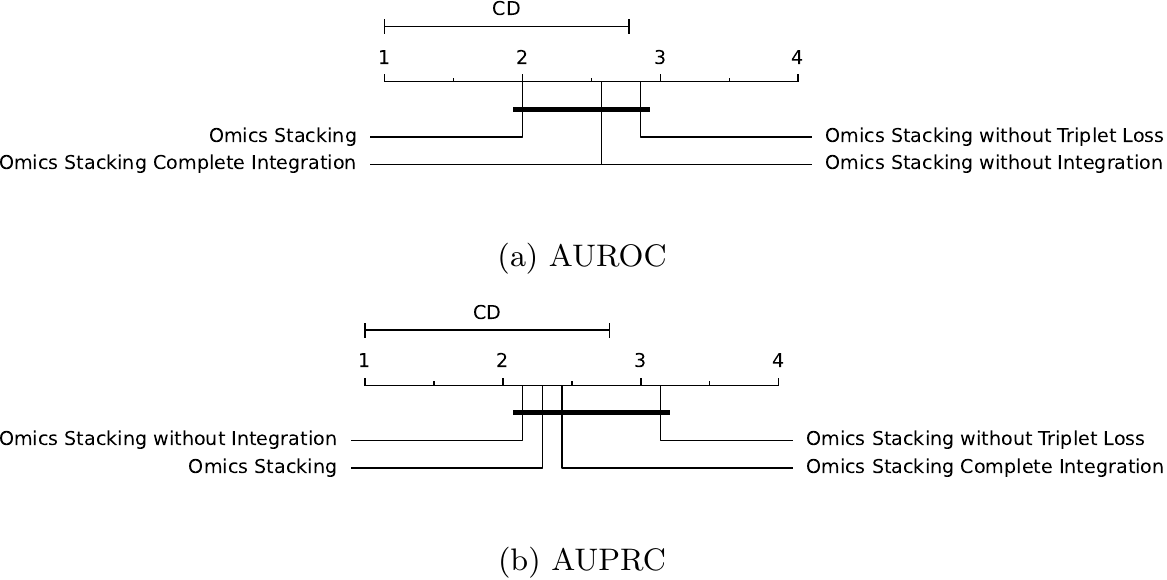}
        \caption{Mean rank and critical difference of the AUROC and AUPRC on the external test data for the ablation study. The mean values of the outer cross-validation results are compared. The Nemenyi test with $\alpha=0.05 $ was used to compute significant differences.}
        \label{fig:critical_difference:external_ablation}
\end{figure}

The results indicate that the omics integration is an essential part and removing it worsens the prediction. Also, the addition of the integration of two omics makes the results worse. We suppose that too many classifiers restrain the network from identifying relevant patterns.
The  AUROCs, AUPRCs and mean ranks are given in \Cref{tab:ablation_auroc_test,tab:ablation_auprc_test,tab:ablation_auroc_external,tab:ablation_auprc_external} and the visualization of them in  \Cref{fig:critical_difference:test_ablation,fig:critical_difference:external_ablation}.

\begin{table*}[ht]
    \centering
    \caption{Mean AUROC on test sets from cross-validation for the ablation study. Best results are shown in bold and second best are underlined. The values represent the means and standard deviations over five iterations.}
    \resizebox{\textwidth}{!}{%
    \begin{tabular}{lcccccccc}
    \toprule
        & \multicolumn{4}{c}{Omics Stacking}\\
        \cmidrule{2-5}
        Drug & Complete Integration & Without Integration &  Integration & Without Triplet Loss \\
        \midrule
        Gemcitabine TCGA  & \underline{$0.630\pm0.055$} & $0.605\pm0.076$ &  $\mathbf{0.646\pm0.045}$ & $0.601\pm0.084$ & \\
        Gemcitabine PDX  & $0.640\pm0.089$ & $\mathbf{0.656\pm0.062}$ &  \underline{$0.651\pm0.071$} & $0.593\pm0.044$ & \\
        Cisplatin  & \underline{$0.742\pm0.050$} & $\mathbf{0.757\pm0.061}$ &  $0.722\pm0.066$ & $0.734\pm0.091$ & \\
        Docetaxel  & \underline{$0.775\pm0.089$} & $0.759\pm0.031$ &  $0.772\pm0.077$ & $\mathbf{0.813\pm0.024}$ & \\
        Erlotinib  & $0.696\pm0.101$ & \underline{$0.744\pm0.098$} &  $\mathbf{0.754\pm0.114}$ & $0.662\pm0.112$ & \\
        Cetuximab  & $\mathbf{0.748\pm0.048}$ & $0.679\pm0.052$ &  $0.731\pm0.090$ & $0.721\pm0.055$ & \\
        Paclitaxel & $\mathbf{0.695\pm0.104}$ & $0.634\pm0.114$ &  \underline{$0.667\pm0.138$} & $0.522\pm0.085$ & \\
        \midrule
        Rank & \textbf{2.00} & 2.57 & \underline{2.14} & 3.29\\ 
    \bottomrule
    \end{tabular}}
    \label{tab:ablation_auroc_test}
\end{table*}

\begin{table*}[ht]
    \centering
    \caption{Mean AUPRC on test sets from cross-validation for the ablation study. Best results are shown in bold and second best are underlined. The values represent the means and standard deviations over five iterations.}
    \resizebox{\textwidth}{!}{%
    \begin{tabular}{lcccccccc}
    \toprule
    & \multicolumn{4}{c}{Omics Stacking}\\
        \cmidrule{2-5}
        Drug & Complete Integration & Without Integration &  Integration & Without Triplet Loss \\
        \midrule
        Gemcitabine TCGA  & \underline{$0.143\pm0.053$} & $0.136\pm0.054$ & $\mathbf{0.161\pm0.053}$ & $0.149\pm0.074$  \\
        Gemcitabine PDX  & $0.138\pm0.046$ & $\mathbf{0.204\pm0.092}$ & \underline{$0.151\pm0.085$} & $0.119\pm0.046$  \\
        Cisplatin  & $0.288\pm0.072$ & $\mathbf{0.293\pm0.074}$ & $\mathbf{0.293\pm0.089}$ & $0.277\pm0.084$    \\
        Docetaxel  & \underline{$0.317\pm0.064$} & $0.262\pm0.072$ & $0.316\pm0.083$ & $\mathbf{0.358\pm0.046}$    \\
        Erlotinib  & $0.422\pm0.704$ & \underline{$0.440\pm0.149$} & $\mathbf{0.479\pm0.153}$ & $0.352\pm0.111$   \\
        Cetuximab  & $\mathbf{0.379\pm0.102}$ & $0.264\pm0.042$ & \underline{$0.376\pm0.104$} & $0.350\pm0.049$ \\
        Paclitaxel & $0.186\pm0.076$ & \underline{$0.188\pm0.103$} & $\mathbf{0.220\pm0.108}$ & $0.090\pm0.024$  \\
        \midrule
        Rank & \underline{2.57} & \underline{2.57} & \textbf{1.71} & 3.14 \\
    \bottomrule
    \end{tabular}}
    \label{tab:ablation_auprc_test}
\end{table*}

\begin{table*}[h]
    \centering
    \caption{Mean AUROC on external test set for the ablation study. Best results are shown in bold and second best are underlined.  The values represent the means and standard deviations over five iterations.}
    \resizebox{\textwidth}{!}{%
    \begin{tabular}{lcccccccc}
    \toprule
    & \multicolumn{4}{c}{Omics Stacking}\\
        \cmidrule{2-5}
        Drug & Complete Integration & Without Integration &  Integration & Without Triplet Loss \\
        \midrule
        Gemcitabine TCGA  & \underline{$0.646\pm0.048$} & $0.641\pm0.056$ & $\mathbf{0.655\pm0.029}$ & $0.624\pm0.068$   \\
        Gemcitabine PDX  & $0.656\pm0.049$ & $0.630\pm0.066$ & $\mathbf{0.714\pm0.089}$ & \underline{$0.665\pm0.078$}  \\
        Cisplatin  & \underline{$0.668\pm0.046$} & $\mathbf{0.685\pm0.078}$ & $0.644\pm0.087$ & $0.680\pm0.063$   \\
        Docetaxel  & $\mathbf{0.613\pm0.058}$ & $0.597\pm0.046$ & $0.584\pm0.101$ & \underline{$0.600\pm0.090$}  \\
        Erlotinib  & $0.704\pm0.141$ & $0.715\pm0.193$ & $\mathbf{0.744\pm0.065}$ & \underline{$0.741\pm0.077$}  \\
        Cetuximab  & $0.529\pm0.052$ & $\mathbf{0.599\pm0.173}$ & \underline{$0.575\pm0.049$} & $0.466\pm0.122$ \\
        Paclitaxel & \underline{$0.511\pm0.106$} & $0.432\pm0.087$ & $\mathbf{0.619\pm0.152}$ & $0.419\pm0.082$ \\
        \midrule
        Rank & \underline{2.57} & \underline{2.57} & \textbf{2.00} & 2.86 \\ 
    \bottomrule
    \end{tabular}}
    \label{tab:ablation_auroc_external}
\end{table*}

\begin{table*}[h]
    \centering
    \caption{Mean AUPRC on the test sets from cross-validation for the ablation study. Best results are shown in bold and second best are underlined.  The values represent the means and standards deviation over five iterations.}
    \resizebox{\textwidth}{!}{%
    \begin{tabular}{lcccccccc}
    \toprule
    & \multicolumn{4}{c}{Omics Stacking}\\
        \cmidrule{2-5}
        Drug & Complete Integration & Without Integration &  Integration & Without Triplet Loss \\
        \midrule
        Gemcitabine TCGA  & $0.537\pm0.040$ & \underline{$0.545\pm0.088$} & $\mathbf{0.581\pm0.074}$ & $0.489\pm0.070$ \\
        Gemcitabine PDX  & $\mathbf{0.522\pm0.087}$ & $0.455\pm0.072$ & \underline{$0.510\pm0.132$} & $0.483\pm0.109$ \\
        Cisplatin  & \underline{$0.954\pm0.009$} & $0.953\pm0.022$ & $0.942\pm0.027$ & $\mathbf{0.955\pm0.012}$ \\
        Docetaxel  & \underline{$0.571\pm0.028$} & $\mathbf{0.644\pm0.048}$ & $0.560\pm0.051$ & $0.568\pm0.056$ \\
        Erlotinib  & $0.400\pm0.171$ & $\mathbf{0.482\pm0.251}$ & \underline{$0.440\pm0.075$} & $0.421\pm0.126$ \\
        Cetuximab  & $0.107\pm0.012$ & $\mathbf{0.188\pm0.164}$ & \underline{$0.125\pm0.018$} & $0.102\pm0.024$ \\
        Paclitaxel & \underline{$0.176\pm0.094$} & $0.127\pm0.027$ & $\mathbf{0.256\pm0.147}$ & $0.121\pm0.017$ \\
        \midrule
        Rank & 2.43 & \textbf{2.14} & \underline{2.29} & 3.14 \\ 
    \bottomrule
    \end{tabular}}
    \label{tab:ablation_auprc_external}
\end{table*}

\end{document}